\documentclass[lettersize,journal]{IEEEtran}
\usepackage{amsmath,amsfonts}
\usepackage{algorithmic}
\usepackage{algorithm}
\usepackage{array}
\usepackage[caption=false,font=normalsize,labelfont=sf,textfont=sf]{subfig}
\usepackage{textcomp}
\usepackage{stfloats}
\usepackage{url}
\usepackage{verbatim}
\usepackage{graphicx}
\usepackage{cite}
\usepackage{multirow}
\usepackage{orcidlink}
\usepackage[table]{xcolor}  
\definecolor{lightgray}{gray}{0.9}  

\begin{document}

\title{A Framework for Real-Time Volcano-Seismic Event Recognition Based on Multi-Station Seismograms and Semantic Segmentation Models\thanks{\textbf{This manuscript is under revision by \textit{IEEE Transactions on Geoscience and Remote Sensing}.}}}

\author{Camilo Espinosa-Curilem~(ORCID: 0009-0003-8243-7854), 
        Millaray Curilem~\IEEEmembership{Member, IEEE Computational Intelligence Society}~(ORCID: 0000-0001-8685-221X), 
        and Daniel Basualto~(ORCID: 0000-0001-7979-2415)%
\thanks{Camilo Espinosa-Curilem is with the Advanced Mining Technologies Center, Universidad de Chile, Santiago, Chile, and the CIVUR Project N°FRO2193/CIV23-0001, Universidad de La Frontera, Temuco, Chile (email: camilo.espinosa@ug.uchile.cl).}%
\thanks{Millaray Curilem is with the Dept. of Ingeniería Eléctrica, Universidad de La Frontera, Temuco, Chile (email: millaray.curilem@ufrontera.cl).}%
\thanks{Daniel Basualto is with the CIVUR Project N°FRO2193/CIV23-0001, the Network for Extreme Environmental Research (NEXER), and Laboratorio Natural Andes del Sur de Chile, Universidad de La Frontera, Temuco, Chile (email: daniel.basualto@ufrontera.cl).}%
}

\markboth{.}{IEEEtran \LaTeX \ Template}

\IEEEpubid{0000--0000/00}

\maketitle

\begin{abstract}
In volcano monitoring, effective recognition of seismic events is essential for understanding volcanic activity and raising timely warning alerts. Traditional approaches rely on manual analysis, which is inherently subjective and resource-intensive, as continuous monitoring necessitates the uninterrupted presence of analysts. Furthermore, current automatic approaches typically address either detection or classification in isolation and predominantly rely on single-station data, limiting their robustness, adaptability, and real operational implementation. We introduce an end-to-end framework that uses a transformation of 1-D multi-station waveforms into compact 2-D image patches. This representation allows deep semantic-segmentation models to generate per-pixel softmax activation maps that simultaneously detect and label events within a fixed time window, while also supporting a sliding-window setup for continuous data streams. We evaluated five architectures (UNet, UNet++, DeepLabV3+, SwinUNet, and PhaseNet) using 24,493 labeled windows from four Chilean volcanoes spanning five event classes, as well as a 10-hour continuous trace. All 2-D models demonstrated strong noise robustness and adaptability to new datasets. UNet achieved the best performance, with a mean F1-score of 0.91 and IoU of 0.88 on labeled windows, and a mean F1-score of 0.68 when applied to the continuous trace. Depending on the architecture and refresh rate, the system can process a 10-hour trace in as little as 4 seconds and up to 5 minutes on an NVIDIA RTX 3060 GPU. We believe the proposed approach provides a practical and effective solution for continuous, real-time volcanic surveillance, as it minimizes preprocessing, leverages spatial correlations across stations, and successfully adapts to unseen volcanoes.
\end{abstract}


\section{Introduction}
\IEEEPARstart{S}{eismic} data is the most widely used and reliable method for monitoring volcanic activity \cite{Saccorotti2021, Carniel2021}. Continuous seismic signals from multiple stations around a volcano are recorded and analyzed to identify patterns that may indicate volcanic processes or events. Traditionally, analysts manually examine the signals to detect seismic events and interpret them. However, this approach is subjective, labor-intensive, and becomes impractical as the number of monitored volcanoes increases or during periods of heightened volcanic activity. Consequently, numerous efforts have been made to develop automatic event detectors that assist human analysts.

A prominent approach in automatic volcano monitoring is the application of machine learning techniques, including Cluster Analysis \cite{Ren2020}, Support Vector Machines (SVM) \cite{Masotti2006}, Hidden Markov Models (HMM) \cite{Beyreuther2008, cortes2009, Corts2019}, and Deep Learning \cite{Scarpetta2005, Curilem2009, 2019TItos, 2020Titos, Canrio2020, 2020salazar, Martnez2021, Lara2021, Ferreira2023}. These methods have demonstrated good performance compared to traditional manual techniques, but they face some challenges when applied to real-time monitoring.

First, many models focus solely on classification tasks, operating on pre-segmented windows. However, in real-world scenarios, the precise onset and termination of volcanic events are difficult to determine, and models that lack detection capabilities can struggle with real-time monitoring. Although several studies have addressed detection in volcano-seismic data by leveraging advanced neural architectures and engineered features (for example, \cite{Lara-Cueva2016,Bueno2019,Bueno2022,Lara2021,Corts2021} use convolutional or Bayesian networks with carefully selected representations) we found works more closely aligned with the present study in the tectonic seismology literature. In particular, the semantic segmentation models SCALODEEP \cite{Saad2021}, PhaseNet \cite{phasenet} and EQTransformer \cite{Mousavi2020} detect earthquakes and pick P/S arrivals directly from continuous streams, demonstrating promising results in real-time seismic analysis. Despite their success, there are still very few studies applying this type of model to volcano-seismic data. For example, \cite{Zhong2024} evaluated PhaseNet and EQTransformer on a large multi-volcano dataset but focused exclusively on two types of events, assessing only classification accuracy without addressing real-time applicability. This limited scope highlights a clear opportunity to apply this type of models in the volcano-seismic domain, where high signal variability, emergent onsets, and overlapping events present distinct challenges not yet fully addressed by existing work.\IEEEpubidadjcol

Second, most approaches focus on recognition using single-station data, which lacks the robustness that multi-station analysis provides \cite{Ferreira2023, Curilem2016}. Analysts, for instance, typically review data from multiple stations to cross-reference signals and distinguish surface events—detected by only a few stations—from deeper events, which are recorded by many stations \cite{Battaglia2003}. 

Our approach addresses both detection and classification tasks by applying semantic segmentation models to a novel multi-station seismic signal representation. Unlike previous methods that use multichannel 1D inputs \cite{phasenet, Mousavi2020}, we introduce a \emph{stacking} procedure that vertically combines small multichannel patches into a 2D image. This representation follows the spirit of prior models like PhaseNet and EQTransformer, which apply segmentation to multichannel 1-D signals. Our patch-stacking step reorganizes multi-station traces into a 2D layout, preserving both temporal and inter-station structure while allowing the use of well-established 2D segmentation architectures.

The proposed models are used in experiments involving a dataset of 24,493 labeled windows containing events from five categories: Volcano-Tectonic (VT), Tremor (TR), Long-Period (LP), Avalanches (AV), and Ice-Quakes (IQ). These events were recorded at four Chilean volcanoes: Nevados del Chillán Volcanic Complex (NChVC), Laguna del Maule (LDM), Villarrica (VCA), and Puyehue-Cordón Caulle (CAU). Additionally, we assess model performance on a 10-hour continuous trace containing 205 cataloged events. In this setting, models are applied in a sliding window fashion over the continuous data.

On the labeled windows, the models achieve strong detection and classification performance, demonstrating robustness to noise and generalizability across volcanoes. When applied to continuous data, the models show a decline in performance, specially affected by false positives in noise-only segments, a limitation discussed further in the manuscript. Overall, our approach offers a novel and practical deep learning solution for real-time volcano monitoring, enhancing the reliability and efficiency of seismic event recognition.

\section{Proposed Framework}
\subsection{Database}\label{section:database}

Volcanoes consist of intricate networks of chambers and conduits through which magma and gases move, and various processes within the volcanic system lead to distinct seismic patterns \cite{Basualto_2023}, which are recorded by seismic stations. Additionally, external factors unrelated to volcanic activity can generate different seismic signatures, as described by \cite{Wassermann, CANARIO2020}. For this work, five types of events were considered, which comprise three volcano-seismic events: VT events, which are associated with the fracturing of rocks within the volcanic conduits; LP events, that result from sudden movements of magmatic or hydrothermal fluids; TR events, which are caused by sustained pressure disturbances of magmatic or hydrothermal fluids and can be continuous or manifest as a sequence of transient signals similar to LP events; and two non-volcanic events: AV, that occur when masses of snow, ice, or volcanic debris move rapidly down the slopes of the volcano; and IC events generated by the sudden fracturing of ice masses, often linked to glacial movements. An additional Background (BG) class was considered to represent the background noise of the seismograms, that is, when no event is present.

\begin{figure}[H]
\centering
\includegraphics[width=\linewidth]{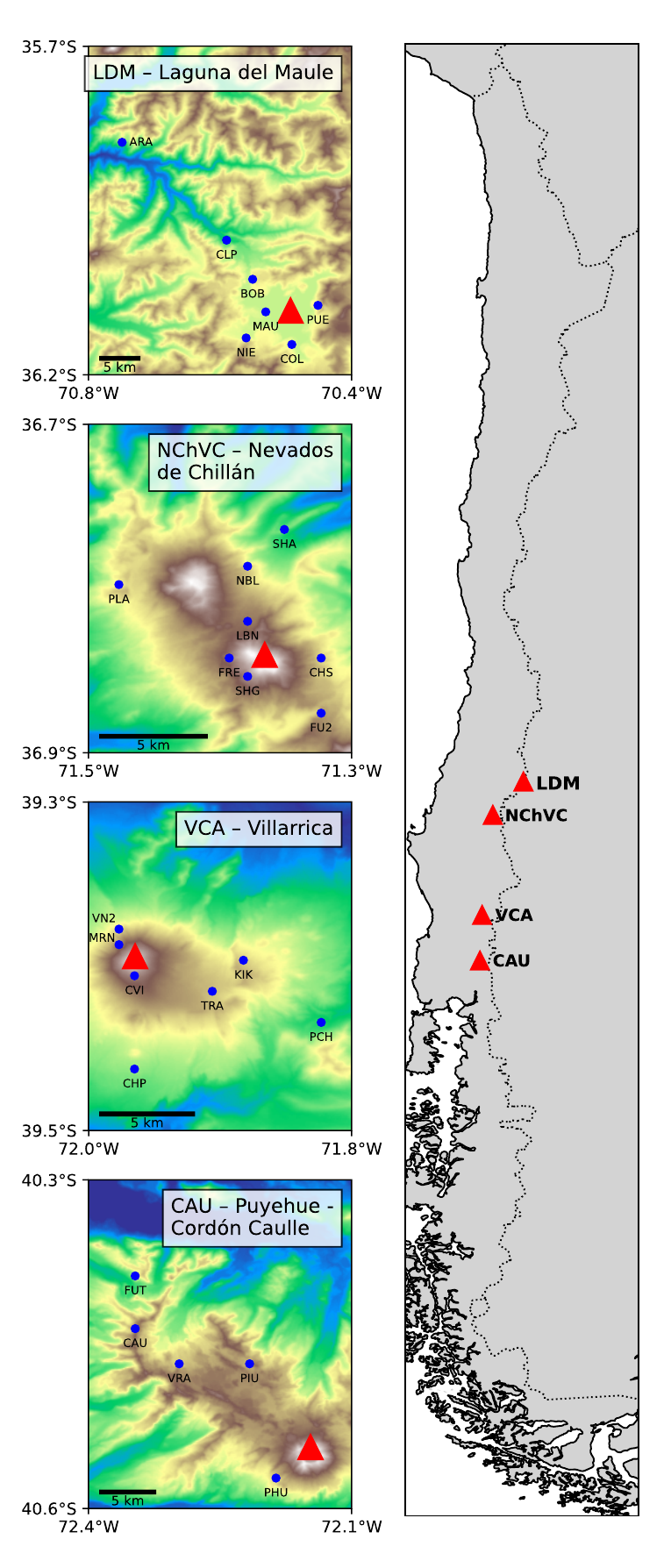}
\caption{Geographical distribution of seismic stations used in this study. 
Left column: detailed station networks around each volcano, ordered from north to south (LDM, NChVC, VCA, and CAU). 
Right column: overview map of Chile showing the relative positions of the four volcanoes. 
Red triangles mark volcano locations, and blue circles indicate seismic stations.}
\label{fig:method_2_stations}
\end{figure}

To train and evaluate our approach, we considered data from four different Chilean volcanoes. We first fit our models on NChVC, and then evaluate the models' scalability when applied to unseen data from three other volcanoes: VCA, LDM, and CAU.

NChVC, located in central Chile, is one of the most active volcanoes in the country. It is an andesitic stratovolcano characterized by frequent vulcanian eruptions, with seismicity dominated by VT, TR, and LP events \cite{Gonzalez-Ferran1995VolcanesChile}. 
The most recent eruptive cycle began in January 2016 and lasted until December 2022 \cite{ovdas2022rav}. This period was marked by vulcanian eruptions, the formation of lava domes, and lava flows \cite{Cardona2021Volcanic2020, Astort2022VolcanicDInSAR}. VCA, a basaltic-andesitic stratovolcano \cite{Corts2024}, has continuous strombolian activity, with VT events being the most common, particularly after eruptions like that of 2015. LDM is a rhyolitic volcanic field with a history of rapid inflation, mostly showing VT events due to magma accumulation \cite{LeMvel2021,Cardona2018}. Finally, the CAU rhyolitic complex is known for fissure eruptions, producing VT, TR, and LP events during major eruptions like the 2011 event \cite{Basualto_2023}.

All seismic stations used in this study operated at a sampling rate of 100 Hz and were equipped with broadband seismometers from one of the following five types: Guralp CMG-6TD, Nanometrics Trillium 120P, Nanometrics Trillium 40P, Nanometrics Trillium COMPACT, and Reftek 151-30. The mean inter-station distances were as follows: NChVC (4.85$\pm$1.01 km, 8 stations), VCA (15.74$\pm$5.15 km, 7 stations), LDM (13.08$\pm$3.50 km, 7 stations), and CAU (11.17$\pm$4.04 km, 5 stations), computed as Cartesian distances in the horizontal (x–y) plane. The seismic stations measured ground velocity, and the instrumental response was removed by Observatorio Vulcanológico Andes Sur (OVDAS)\footnote{\url{https://rnvv.sernageomin.cl/observatorio-volcanologico-de-los-andes-del-sur/}} using a deconvolution procedure based on the sensor’s data sheet parameters (poles, zeros, gain, and sensitivity). This processing ensures accurate velocity representation and eliminates instrument-induced distortions, enabling homogeneous and unbiased analysis. The spatial arrangement of the seismic networks is shown in Figure~\ref{fig:method_2_stations}, with detailed station maps for each volcano (left panels) and an overview map of Chile locating the four volcanic systems (right panel).

Seismic events were randomly sampled for each volcano at different periods of time: NChVC from January 2017 to December 2022, LDM from April 2012 to July 2023, VCA from September 2012 to June 2023, and CAU from January to December 2011. For each event, Z-component seismograms were collected from one to eight stations, depending on availability during the selected period. Unavailable stations were represented as zero-valued arrays. On average, the number of active stations per event was 4.6 for NChVC, 2 for VCA, 5.3 for LDM, and 3.5 for CAU.

To ensure a high-quality dataset for model training, event labeling was provided by OVDAS and reviewed by our team’s volcano seismologist. Each event is manually assigned a single onset and termination time, defined respectively as the first clear arrival and the last visible event-related signal across the network. These times are applied consistently across all stations, ensuring uniform labeling for evaluation. Relevant characteristics of the data are summarized in Table \ref{BD_tab}

\begin{table}[]
\setlength{\tabcolsep}{2.5pt}  
\renewcommand{\arraystretch}{1.2}  
\caption{Summary of duration and frequency characteristics for different volcanic event classes. The frequency range corresponds to the lower and upper frequency limits containing 95\% of the total spectral energy, providing a measure of the typical frequency range for each event class.}
\label{BD_tab}%
\begin{tabular}{cc|c|rclcl}
\hline
\hline
\multicolumn{1}{c|}{\multirow{2}{*}{\textbf{Volcano}}} & \multirow{2}{*}{\textbf{Class}} & \multirow{2}{*}{\textbf{\begin{tabular}[c]{@{}c@{}}Number of \\ Events\end{tabular}}} & \multicolumn{3}{c|}{\textbf{Duration (s)}}                                           & \multicolumn{2}{c}{\textbf{Frequency (Hz)}} \\ \cline{4-8} 
\multicolumn{1}{c|}{}                                  &                                 &                                                                                       & \multicolumn{1}{c}{\textbf{Min}} & \textbf{Mean} & \multicolumn{1}{c|}{\textbf{Max}} & \textbf{Mean Peak}     & \textbf{Range}     \\ \hline
\multicolumn{1}{c|}{\multirow{5}{*}{NChVC}}            & VT                              & 3068                                                                                  & 7.5                              & 16.6          & \multicolumn{1}{l|}{32.1}         & 5.7                    & [1.9 - 12.1]       \\
\multicolumn{1}{c|}{}                                  & LP                              & 1892                                                                                  & 10.5                             & 27.6          & \multicolumn{1}{l|}{59.0}         & 2.6                    & [1.4 - 6.5]        \\
\multicolumn{1}{c|}{}                                  & TR                              & 2360                                                                                  & 75.9                             & 143.3         & \multicolumn{1}{l|}{279.2}        & 2.2                    & [1.3 - 5.9]        \\
\multicolumn{1}{c|}{}                                  & AV                              & 805                                                                                   & 11.0                             & 31.9          & \multicolumn{1}{l|}{83.2}         & 5.4                    & [2.3 - 11.7]       \\
\multicolumn{1}{c|}{}                                  & IC                              & 977                                                                                   & 3.1                              & 7.5           & \multicolumn{1}{l|}{15.2}         & 9.1                    & [3.8 - 14.1]       \\ \hline
\multicolumn{1}{c|}{\multirow{3}{*}{CAU}}              & VT                              & 2298                                                                                  & 8.1                              & 12.1          & \multicolumn{1}{l|}{47.2}         & 3.9                    & [1.2 - 12.5]         \\
\multicolumn{1}{c|}{}                                  & LP                              & 2081                                                                                  & 15.2                             & 28.6          & \multicolumn{1}{l|}{79.4}         & 1.8                    & [1.0 - 11.3]       \\
\multicolumn{1}{c|}{}                                  & TR                              & 2833                                                                                  & 59.3                             & 157.8         & \multicolumn{1}{l|}{935.1}        & 3.1                    & [1.0 - 11.7]       \\ \hline
\multicolumn{1}{c|}{VCA}                               & VT                              & 1516                                                                                  & 8.1                              & 12.0          & \multicolumn{1}{l|}{65.2}         & 3.3                    & [1.2 - 13.0]       \\ \hline
\multicolumn{1}{c|}{LDM}                               & VT                              & 6663                                                                                  & 6.9                              & 14.2          & \multicolumn{1}{l|}{87.1}         & 9.6                    & [2.6 - 14.7]       \\ \hline
\multicolumn{2}{c|}{\textbf{TOTAL}}                                                      & \textbf{24493}                                                                        &                                  &               &                                   &                        &                   
\end{tabular}
\end{table}

To generate the datasets, we extracted fixed-length windows from continuous seismograms at all available stations, selecting only those containing a single event or consecutive parts of one if it was longer. This ensured each window corresponded to a single class, simplifying training and evaluation. The window size reflects a trade-off between computational and data constraints: longer windows demand more resources and are less likely to isolate single events, however, they also offer more context for the models and can accommodate longer events. This trade-off is further addressed in the Discussion section.

To investigate the impact of window size on model performance, datasets were generated for three different window lengths: 80, 20, and 5 seconds. This comparison helps understand the types of information the models consider, particularly regarding frequency content and event duration. An 80-second window captures most events (except for TR), a 20-second window fully contains most VT and IC events, and some LP events, while a 5-second window only completely covers some IC events.

The choice of window size is subject to a soft constraint imposed by Equation~\ref{eqn:window_size} to enable a 2D representation of the signals, as detailed in Section~\ref{section:methods_1}. Window lengths of 80, 20, and 5 seconds were chosen, slightly below the theoretical upper bounds, to illustrate that the folding constraint is flexible and signals can be zero-padded if required.

\subsubsection{Data Preparation}
\label{sec:data_prep}
Two preprocessing steps were applied to the multi-station events in the database: (a) a Butterworth bandpass filter was utilized to filter the events within the frequency range of 1 to 15 Hz, with a filter order of 5 and applied as a zero-phase two-pass filter (using the \texttt{sosfiltfilt} function from the SciPy library), and (b) the signals were normalized by dividing all samples by the maximum absolute value across the stations for each event, ensuring the signal amplitudes are in the range [-1,1].

For the fitting to the NChVC volcano stage, we randomly sampled 200 events for validation and testing from each class. The remaining events were either sampled (for VT, LP and TR) or augmented to reach a total of 1,500 events in the training set (see Table \ref{tab:train_val_test}). Data augmentation over the training dataset consisted in copying random events and randomly shuffling both the order of their stations and the time they started inside the window. This allowed to create station-agnostic models and to balance the dataset by increasing the number of events in underrepresented classes, mainly AV and IC.

\begin{table}
\centering
\caption{Number of events from NChVC used in the training/evaluation of the models.}\label{tab:train_val_test}%
\begin{tabular}{@{}cccc@{}}
\hline
\textbf{Class} & \textbf{Training set} & \textbf{Validation Set} &\textbf{Test Set}\\
\hline
VT  & 1500 & 200 & 200\\
LP  & 1500  & 200 & 200\\
TR  & 1500  & 200 & 200\\
AV  & 1500  & 200 & 200\\
IC  & 1500  & 200 & 200\\
\textbf{TOTAL}  & \textbf{7500} & \textbf{1000} &\textbf{1000} \\
\hline
\end{tabular}
\end{table}

\subsection{Semantic segmentation of multi-channel signals}\label{section:methods_1}

Semantic segmentation is a computer vision task that assigns a class label to each pixel in an image. Early approaches relied on handcrafted features and traditional image processing techniques like thresholding and clustering, but these methods were limited in capturing complex patterns. The introduction of deep learning, particularly convolutional neural networks (CNNs), revolutionized this field. A major breakthrough was the Fully Convolutional Network (FCN) \cite{2017shelhamer}, which enabled pixel-wise classification using CNNs. This paved the way for models like U-Net \cite{2015ronnenberg}, which introduced skip connections to improve accuracy and inspired numerous variations. More recently, attention mechanisms and Visual Transformers \cite{2020Dosovitskiy} have emerged to model global dependencies more effectively, albeit with increased computational and data requirements.

To harness image‑segmentation models for seismic analysis we convert each fixed‑length, multi‑station waveform window into a two‑dimensional picture, pass it through the model, and then project the pixel‑level activations back onto the original one‑dimensional (1‑D) traces.  The forward transform is called \emph{Patch Stacking}; the inverse transform is \emph{Activation Unstacking}. Together they allow us to apply off‑the‑shelf computer‑vision models to raw seismic data with only a few lines of preprocessing code. In the next section we explain how these transforms are applied and illustrative code examples for all the operations are provided in the companion repository (see Section~\ref{section:code_availability}).

\subsubsection{Seismograms as 2‑D images}
\label{sec:patch_stacking}

Let \(\mathbf{X}\in[-1,1]^{S\times W}\) be a normalised window where \(S\) is the number of stations and \(W\) the window length in samples. The \emph{Patch Stacking} procedure slices \(\mathbf{X}\) into \(K=W/N\) contiguous blocks of size \(S\times N\) and places those blocks one beneath the other to form a grayscale image whose side length \(N\) obeys

\begin{equation}
\label{eqn:window_size}
N = \sqrt{S\,W}
\end{equation}

For example, an \(8\times8192\) window (\(S=8,\,W=8192\)) yields \(K=32\) patches of \(8\times256\), which stack into a \(256\times256\) image. Note that the square requirement is only needed by architectures that expect equal spatial dimensions; models such as U‑Net, U-Net++ or DeepLab can accept arbitrary rectangles. Additionally, windows that do not meet the specific $W$ size (e.g. 8000 instead of 8192) can be zero padded without detrimental effects on performance. 

\subsubsection*{Activation Unstacking}

After segmentation the network produces a cube of soft‑max activations with the same spatial dimensions as the input image and \(C=6\) class channels.  
Activation Unstacking concatenates the activation tiles horizontally, reconstructing a tensor of shape \(S\times W\times C\) (e.g. 8 stations $\times$ 8192 samples $\times$ 6 classes).  

\subsubsection{Window--level detection}
\label{sec:window_detection}

Activation Unstacking returns per--sample soft‑max activations \(\mathbf{A}\in[0,1]^{S\times W\times C}\), where \(S\) is the number of stations, \(W\) the window length in samples, and \(C=6\) the number of classes (background (BG) plus five event types). To obtain time-based detection, the activations are first summed over the station axis,
\[
\tilde{\mathbf{A}}_{c}(t)=\sum_{s=1}^{S} \mathbf{A}_{s,t,c},
\qquad c\in\{1,\dots,6\},
\]
yielding six \((1\times W)\) class‑specific traces. At each sample \(t\) the class with the largest activation is set to one and all others to zero (binarization), producing a mutually exclusive, time‑based segmentation. An event starts when the BG trace switches from one to zero and ends when it returns to one. The event label is chosen as the class that occupies the highest fraction of samples inside the segment. Note that this setup supports station-level and simultaneous-event detection if the station-wise summation and binarization procedures are omitted.

\subsubsection{Sliding‑window detection}
\label{sec:sliding_detection}

\begin{figure*}[t]
\centering
\includegraphics[width=\textwidth]{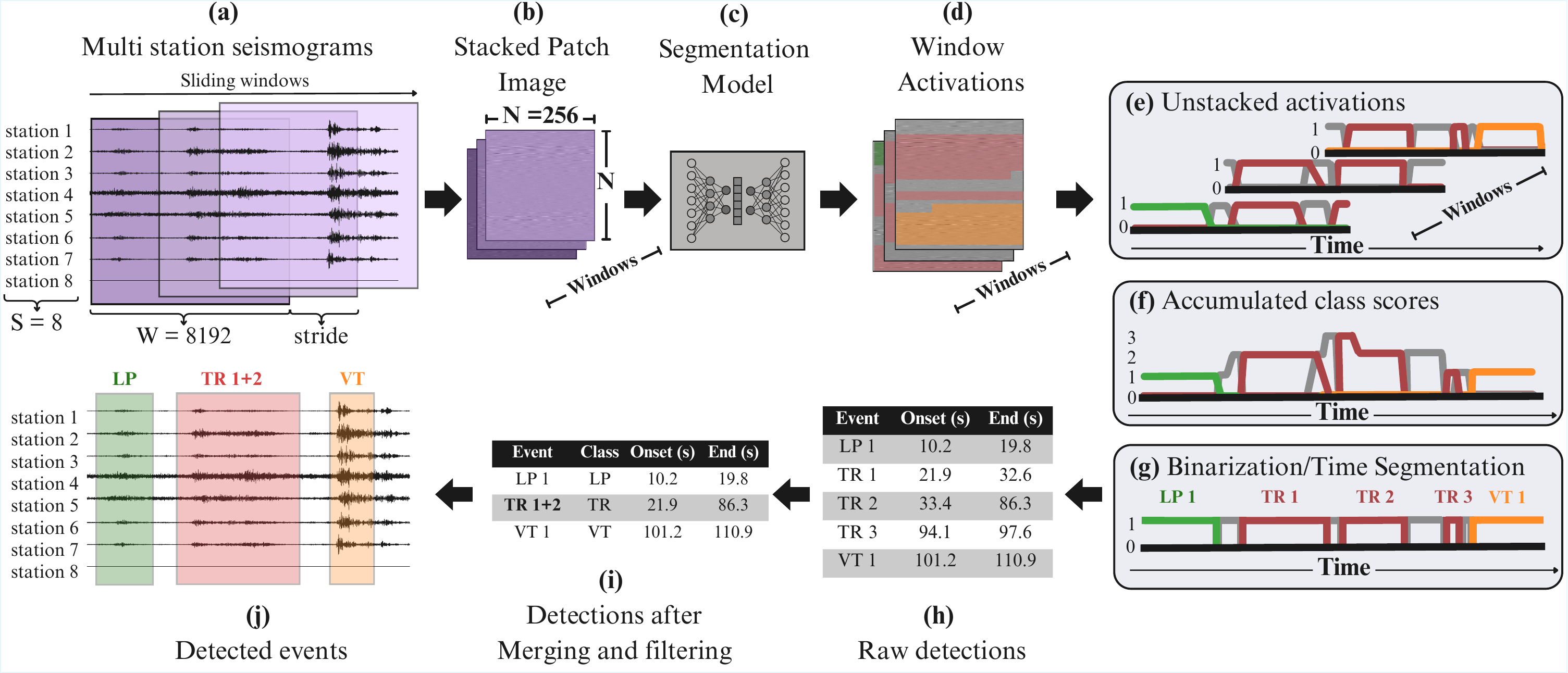}
\caption{Real‑time event‑recognition pipeline for a continuous, multi‑station stream. In the figure $S=8$, $W=8192$ and $N=256$ represent the parameters used for most of the experiments in this work.
(a) The incoming trace is sliced into overlapping windows of \(W\) samples and \(S\) channels. 
(b) Each window is reshaped into a square grayscale image by \emph{Patch Stacking}. 
A 2‑D segmentation model (c) assigns per‑pixel class probabilities to each stacked window (d). 
(e) \emph{Activation Unstacking} and station summation produces window-level activations of shape $C\times W$ ($C=6$ classes). 
(f) Activations are accumulated in a time‑aligned buffer. 
(g) The buffer is binarized to yield a mutually exclusive, time‑based segmentation. 
(h) Detected events are written to a dataframe with assigned class and start/end times. 
(i) A post‑processor merges adjacent segments of the same class separated by less than a user-defined threshold (in our study, 2.5 seconds; e.g. TR 1 and TR 2) and filters out events with unrealistic durations (e.g TR 3 is too short).
The final dataframe records Unix timestamps (or sample indices) for the onset and end of each detected event and the predictions can be visualized over the original data (j). 
For clarity, start and end times are shown in seconds.}
\label{fig:method_0_diagram}
\end{figure*}

For continuous streams, the model operates in a sliding‐window fashion. Each step yields an \emph{unstacked} activation tensor, which is added into a running buffer aligned to absolute time. In this step, each window’s activations are written into the buffer at their corresponding time indices, and when windows overlap their activations are summed. This produces a continuous, time-aligned activation map (Figure~\ref{fig:method_0_diagram}f), which is then processed exactly as in Section~\ref{sec:window_detection}: activations are summed across stations, binarized, and analyzed via BG transitions to determine onset and offset times, with event labels derived from class proportions. 

We sought to minimize reliance on heuristics; however, two straightforward rules were necessary to keep the model outputs consistent with the data. First, \emph{temporal merging} joins segments of the same class that are separated by less than 2.5 s, this is consistent with a local seismic network confined to a stratovolcano, where an event recorded at different stations with a delay of less than 2.5 seconds is generally considered "the same event" for the entire network. Second, \emph{duration filtering} discards any resulting detection whose length falls outside the empirical range observed for its class. The limits, estimated from the 24,493 labeled windows, were \textit{VT}: 6–90 s, \textit{LP}: 10–80 s, \textit{TR}: $\ge$55 s, \textit{AV}: 11–100 s, and \textit{IC}: 3–16 s. 

Figure~\ref{fig:method_0_diagram} summarizes the real‐time pipeline—showing \emph{Patch Stacking}, activation accumulation, segmentation, and post‐processing. In this study, the buffer tensor spans the entire 10-hour trace (a \texttt{.npy} file of approximately 180~MB). We did not explore alternative sizes, as real-time priorities and computational constraints will dictate different requirements for this auxiliary element; nonetheless, we believe our implementation is representative of the intended usage.  

\subsection{Models}

We benchmarked four state‑of‑the‑art architectures for two‑dimensional semantic segmentation—U‑Net, UNet++, DeepLabV3+, and Swin‑UNet—and, for comparison, the one‑dimensional PhaseNet model originally developed for tectonic event detection and phase picking. Each 2‑D network was taken from its reference implementation and then iteratively downsized by pruning filters or encoder–decoder blocks until the first noticeable drop in validation accuracy; the resulting “compact” variants retain full performance while lowering memory demand and reducing the risk of overfitting. PhaseNet, which is already lightweight, was treated in the opposite way: convolutional channels were progressively added until its accuracy plateaued. We aimed to provide an upper bound on what 1‑D segmentation can achieve for volcano‑seismic data and a fair baseline against the 2‑D approaches evaluated in this study.

The original U-Net architecture \cite{2015ronnenberg} was initially developed for biomedical image segmentation. It comprises an encoder-decoder structure with skip connections that preserve spatial information, enabling precise localization while simultaneously capturing contextual information. In our implementation, we set the initial number of features at 16 and a depth of 5 encoder/decoder levels, resulting in a total of 7,778,406 trainable parameters.

An extension of the U-Net model, UNet++ \cite{Zhou2018} introduces nested skip pathways to improve feature propagation and enhance segmentation accuracy. UNet++ refines the segmentation outputs at various levels of the network, capturing fine details and mitigating the vanishing gradient problem, thus achieving better performance in complex segmentation scenarios. We used the EfficientNet-B1 encoder with a depth of 3 encoder/decoder levels and set the decoder channels' sizes to [64, 32, 16], resulting in 7,116,710 trainable parameters.

DeepLabV3+ \cite{Chen2018} builds upon the DeepLabV3 framework \cite{2016Chen} by incorporating a decoder module to refine segmentation results. Unlike U-Net, it employs spatial pyramid pooling and atrous convolutions to capture multi-scale contextual information. This approach is particularly effective for segmenting objects at different scales and has been widely adopted for its robustness and efficiency. In our implementation, we used the MobileNetV2 encoder as backbone with a depth of 5 and a decoder channel size of 128. The total number of trainable parameters was 3,204,694, making it the smallest model we tested.

To leverage the strengths of attention mechanisms, Swin-UNet \cite{2021Cao} replaces convolutional layers with Swin Transformer modules. This enables a more complex and global representation of the inputs. The Swin-Transformer architecture employs shifted-window attention to reduce computational and data requirements compared to other Visual Transformers. In our setup, we used a reduced embedding dimension of 48 (from the original 96), with a window size of 8, a depth of [2, 2, 2, 2] for the encoder and [1, 2, 2, 2] for the decoder. The model uses a reduced number of attention heads ([3, 6, 8, 12]) to balance complexity and performance, resulting in 6,832,164 trainable parameters.

PhaseNet \cite{phasenet} is a U-Net-based 1D convolutional network designed for phase picking and detection tasks in tectonic seismicity. We adapted it for volcanic event recognition to assess the limitations of 1D segmentation models in this context. Our implementation of PhaseNet employs a deep convolutional architecture with a kernel size of 7, a stride of 2, a depth of 5, and a root filter size of 32 (analogous to the initial number of features in U-Net). The normalization scheme used is standardization ("std"). This configuration resulted in 4,279,078 trainable parameters.

Across all implementations, the number of trainable parameters is largely unaffected by input size, since convolutional kernels and layer connections are independent of spatial dimensions. Consequently, U-Net, UNet++, PhaseNet, and DeepLabV3+ preserve a fixed parameter count under different input resolutions, while Swin-UNet shows only very minor variations due to resolution-dependent components in its transformer design. The reported parameter counts can be reproduced using the exploratory material included in our repository.

\subsection{Training setup}
\label{sec:train_setup}

As detailed in Section~\ref{section:database}, the dataset comprises 9\,500 labeled windows: 7\,500 for training, 1\,000 for validation, and 1\,000 for testing. Each target is a patch‑stacked activation tensor of shape \(C\times N\times N\) with \(C=6\) classes and $N=256,128 \text{ or }56$ depending on the window size (see Section~\ref{sec:evaluation}).

All experiments were run in PyTorch 1.13 on a single NVIDIA RTX3060 GPU (6\,GB VRAM). Source codes were taken from the original repositories: U‑Net\footnote{\url{https://github.com/mateuszbuda/brain-segmentation-pytorch}}, UNet++ and DeepLabV3+\footnote{\url{https://github.com/qubvel-org/segmentation_models.pytorch}}, Swin‑UNet\footnote{\url{https://github.com/HuCaoFighting/Swin-Unet}}, and PhaseNet via \texttt{seisbench}\footnote{\url{https://github.com/seisbench/seisbench}}.

Models were trained for 200 epochs with a default AdamW implementation (\(\beta_1\!=\!0.9,\;\beta_2\!=\!0.999,\;\lambda\!=\!10^{-2}\)). The learning rate followed a cosine‑annealing schedule between \(1\times10^{-4}\) and \(1\times10^{-5}\) every 25 epochs. Early stopping terminated training after 30 stagnant epochs for Swin‑UNet and after 10 for the remaining models; the checkpoint with the lowest validation loss was retained.

Because tremor (TR) windows occupy far more pixels than other classes, the data are spatially imbalanced.  Optimization therefore used the Dice loss~\cite{Sudre2017}, which emphasizes region overlap rather than raw pixel count and stabilizes gradients for both rare and dominant classes. 

\subsection{Evaluation}
\label{sec:evaluation}

Our models are expected to both detect the presence of a seismic event and assign it to the correct class. We report Intersection over Union (IoU) \cite{IoU_} for the detection task and F1-score score \cite{F1_score} for the classification task. Metrics are computed in four scenarios that match the intended use cases of the system:

\subsubsection{Evaluation of Window-level Performance}
\label{sec:det_class_eval}

Models are first evaluated on the NChVC test set to assess in-domain performance. Results are reported for three window sizes (80 s, 20 s, and 5 s) to expose any sensitivity to temporal context. Detection performance is quantified with Intersection-over-Union (IoU) between the predicted event mask, obtained by inverting the background channel (event = 1, background = 0), and the likewise inverted reference mask for the same window. Classification is evaluated at the window level: for each window, the model produces a class-probability array with one dimension for the number of classes (five event classes plus background) and one for the number of time samples. We sum the predicted samples per class and assign the window to the class with the largest sum (excluding background). This predicted class is then compared against the ground-truth label to compute the F1-score, which balances precision (fraction of predicted events that are correct) and recall (fraction of true events that are correctly classified). For overall performance, we report the macro averages

\[
\text{F1}_{\mathrm{macro}} = \frac{1}{C}\sum_{i=1}^{C} \text{F1}_{i},
\qquad
\text{IoU}_{\mathrm{macro}} = \frac{1}{C}\sum_{i=1}^{C} \text{IoU}_{i},
\]

where \(C=5\) is the number of classes. Both metrics range from 0 to 1, with higher values indicating better performance. Because the dataset contains only event windows, the background class was excluded from F1 evaluation; background is instead assessed separately through the IoU-based detection metric.

\subsubsection{Evaluation of Noise Robustness}
\label{sec:noise_eval}

\begin{figure}
\centering
\includegraphics[width=0.95\linewidth]{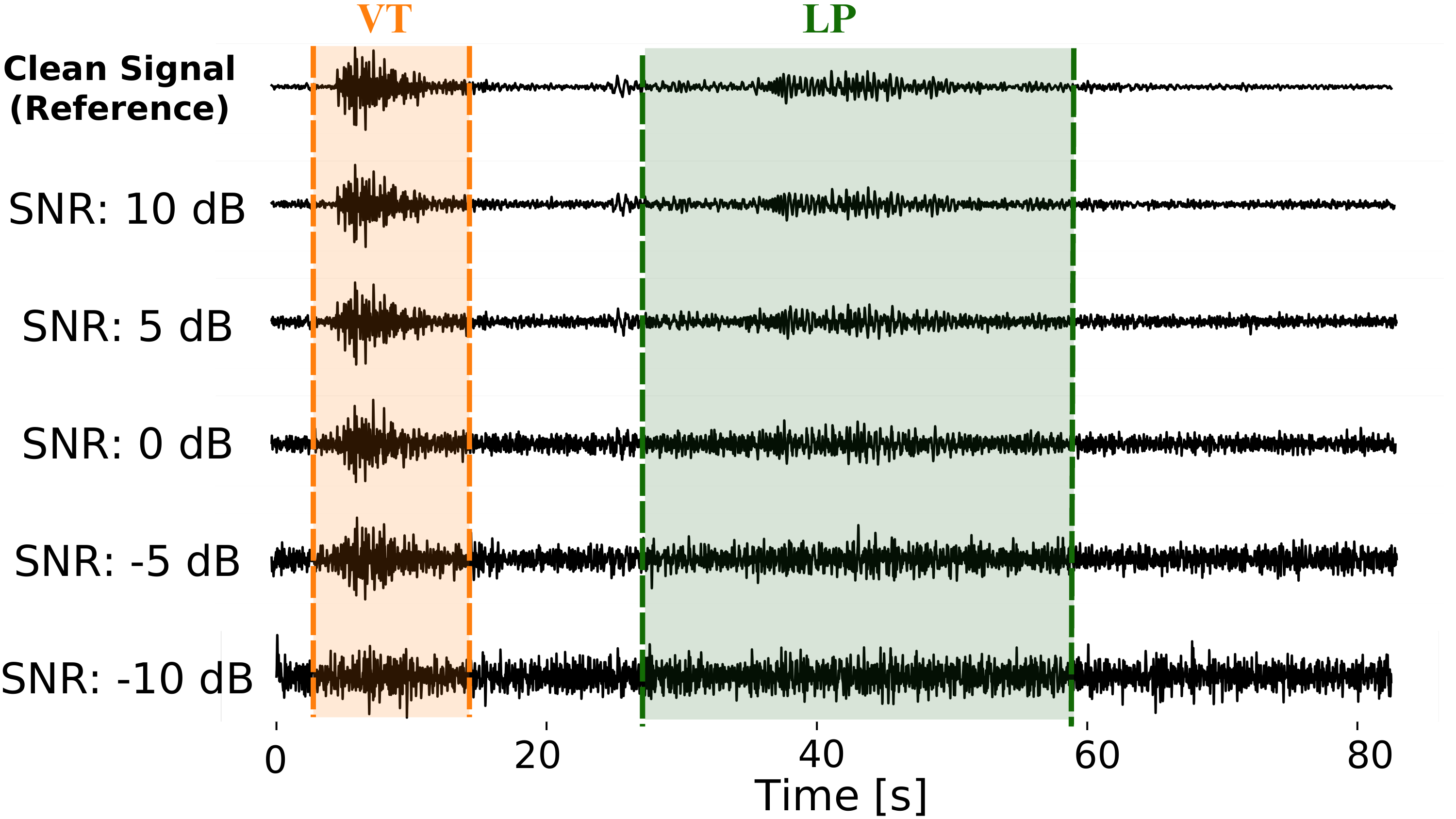}
\caption{Comparison of seismic signals at varying signal-to-noise ratios (SNR) for a single channel of the same seismic window. The top panel shows the clean signal (containing two events) as reference, while subsequent panels illustrate the signal at decreasing SNR levels (10 dB, 5 dB, 0 dB, -5 dB, -10 dB). Y-axis values are omitted for clarity.}
\label{fig:method_1_NOISE_examples}
\end{figure}

Robustness to noise is measured on the NChVC test set with additive white Gaussian noise that is band‑limited to the signal passband \(1\text{--}15\ \text{Hz}\), simulating on‑line conditions after basic filtering. The noise amplitude is scaled to a desired signal‑to‑noise ratio (SNR),
\[
\text{SNR (dB)} = 10 \log_{10}\!\left(\frac{P_{\text{signal}}}{P_{\text{noise}}}\right),
\]
and 21 noisy sets are generated from \(-10\) dB to \(+10\) dB in 1 dB steps. For each SNR we compute mean IoU and mean F1-score to quantify the degradation in accuracy. Figure~\ref{fig:method_1_NOISE_examples} illustrates a single trace at five SNR levels.

\subsubsection{Assessment of Model Flexibility}
\label{sec:flex_eval}

We evaluate two settings on three additional volcano datasets (see Section~\ref{section:database}): a zero-shot regime, where models trained on NChVC are applied without further training; and mixed fine-tuning (transfer learning) with \(1\%\), \(5\%\), \(10\%\), and \(20\%\) of labeled target data, holding out the remaining \(80\%\) for testing. Because the new datasets contain only a subset of the original classes (Table~\ref{BD_tab}), we mitigate the lack of class examples (a known source of catastrophic forgetting) by performing \emph{class completion} within each training split~\cite{Cao2018}. Concretely, for VCA and LDM (VT-only) we add an equal number of randomly sampled NChVC examples per missing class (LP, TR, AV, IC) matching the VT count in that fraction; for CAU (VT/LP/TR present) we add AV and IC from NChVC in a quantity equal to the mean of \{VT, LP, TR\} in that fraction. This design is based on established class-incremental strategies that rebalance old vs. new classes under imbalanced updates~\cite{Hou2019,Mittal2021}. Performance on the held-out test set is reported as IoU (detection), mean F1 (classification on CAU), and recall (classification on VT-only datasets: VCA and LDM). Data splits and class counts are summarized in Table~\ref{tab:new_volcanoes}.

\begin{table}[]
\centering
\caption{Data distribution of VCA, LDM and CAU volcanoes for the evaluation of model flexibility.}\label{tab:new_volcanoes}
\begin{tabular}{c|c|cccc|c}
\multirow{2}{*}{\textbf{Volcano}} & \multirow{2}{*}{\textbf{\begin{tabular}[c]{@{}c@{}}Test Set\\ Size (80\%)\end{tabular}}} & \multicolumn{4}{c|}{\textbf{Train Set Size}} & \multirow{2}{*}{\textbf{Total}} \\ \cline{3-6}
 &  & 1\% & 5\% & 10\% & 20\% &  \\ \hline
VCA & 1212 & 15 & 76 & 152 & 304 & 1516 \\
LDM & 5329 & 66 & 333 & 667 & 1334 & 6663 \\
CAU & 5768 & 71 & 360 & 720 & 1444 & 7212 \\ \hline
\end{tabular}
\end{table}

\subsubsection{Evaluation on Continuous Data}
\label{sec:cont_eval}

The full real--time pipeline in Figure ~\ref{fig:method_0_diagram} is applied over a ten--hour continuous record from NChVC built by concatenating three annotated stretches: a six--hour segment rich in LP, TR, and AV events, a three--hour segment dominated by VT events, and a one--hour segment containing mostly IC events. Sliding windows were processed, activations accumulated, and overlapping detections merged as described in Section~\ref{sec:sliding_detection}.

A prediction was considered correct when at least half of its duration overlapped a reference event. Formally, we use the \emph{Intersection-over-Prediction} (IoP) measure,
\[
    \mathrm{IoP}=\frac{|P \cap R|}{|P|},
\]
where \(P\) and \(R\) denote the predicted and reference time intervals, respectively. IoP differs from the more common Intersection-over-Union in that the denominator contains only the predicted length. This choice is motivated by real-time monitoring practice: analysts are primarily concerned with the proportion of each alert that corresponds to a genuine event, and they generally prefer to avoid penalizing slight under-segmentation that results in short alerts. Operational experience has shown that the most critical task is the correct classification of the event itself; identifying its end (i.e., the coda) is often subject to interpretation and debate. Therefore, discrepancies in defining the end-time window of seismic event do not necessarily affect the classification of the event.

The prediction dataframe is matched to the reference dataframe and mean precision, recall, and F1-score are computed for every model. Results are reported in two complementary variants. In the class-specific setting, a detection is counted only when its predicted label exactly matches the reference event class, thus evaluating the full detection-plus-classification task. In the class-agnostic setting, a detection is credited as long as it overlaps the reference event with IoP $>$ 0.5, irrespective of the predicted label, thereby isolating the models’ raw ability to flag events in the presence of noisy, continuous data.

We evaluate nine inference strides for the sliding window (2.5, 5, 10, 20, 30, 40, 50, 60, 70~s) to test sensitivity to window overlap. The stride determines the model’s \emph{refresh rate} (and thus alert latency): a 2.5~s stride updates predictions at least every 2.5~s, enabling earlier alerts but with greater overlap and computational cost; larger strides reduce overlap and cost at the expense of slower updates.

\begin{table}[]
\centering
\caption{Event counts in the three annotated continuous segments used for
real‑time evaluation.}
\label{tab:real_time_data}
\begin{tabular}{c|ccccc|c}
\hline
Segment             & \textbf{VT} & \textbf{LP} & \textbf{TR} & \textbf{AV} & \textbf{IC} & Total \\ \hline
\textbf{6h segment} & 12           & 50          & 33          & 19          & 13          & 127    \\
\textbf{3h segment} & 24          & 5           & 5           & 2           & 9           & 45    \\
\textbf{1h segment} & 2           & 0           & 0           & 2           & 29          & 33    \\ \hline
\textbf{Total}      & 38          & 55          & 38          & 23          & 51          & 205   \\ \hline
\end{tabular}
\end{table}

Because the models were not trained on noise-only windows, short low-amplitude transients can trigger spurious detections on continuous data. We therefore apply a simple, fixed amplitude threshold based on RSAM, a widely used operational metric in volcano seismology~\cite{Endo1991}.
For station \(k\) and a detected segment \(I\),
\[
\mathrm{RSAM}_k(I)=\frac{1}{|I|}\sum_{t\in I}\!|x_k(t)|\, .
\]

\begin{figure}
\centering
\includegraphics[width=\linewidth]{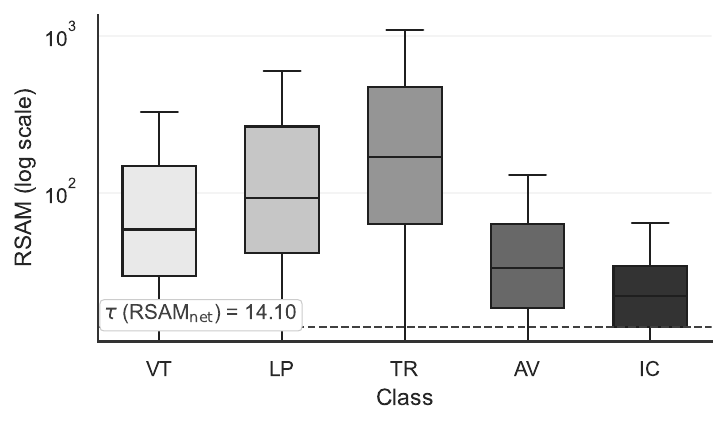}
\caption{Class-specific \(\mathrm{RSAM}_{\text{net}}\) values in the NChVC training dataset, with the dashed line indicating the smallest first-quartile value across the five classes.}
\label{fig:RSAM_per_class}
\end{figure}

In this calculation, we discard zero-filled channels and stations with \(\mathrm{SNR}_k < 0.5\,\mathrm{dB}\), where \(\mathrm{SNR}_k\) is computed between the detected event and a pre-event window of equal length. For each detection, we obtain \(\mathrm{RSAM}_{\text{net}}\) as the median RSAM across all viable stations. The threshold \(\tau\) is derived from the NChVC segmented (training) windows (see Figure~\ref{fig:RSAM_per_class}). Specifically, we take the first quartile of the lowest class (IC in this case), and events with \(\mathrm{RSAM}_{\text{net}}\) below this threshold are discarded.

\section{Results}

\subsection{Data Fitting over NChVC}

Table~\ref{tab:results_1} reports the F1-score and IoU for each architecture at the evaluated window lengths on the NChVC test set. For the largest window (80 s), the four 2-D models—UNet, UNet++, DeepLabV3+, and SwinUNet—perform similarly, with both metrics ranging from 0.86 to 0.91. UNet attains the highest F1-score and IoU performance at 0.91 and 0.88 respectively.  
The 1-D baseline, PhaseNet, performs worse than the 2-D models by a wide margin (F1 = 0.64 and IoU = 0.82).

As the window length decreases, mean F1 steadily declines even though IoU shows a modest rise. We believe the IoU gain merely reflects tighter alignment between shorter windows and event boundaries so overall performance nonetheless deteriorates with smaller windows. Due to this pronounced degradation in performance, we restrict our analysis in the following sections to results obtained using 80-second windows only.

\begin{table*}[]
\caption{F1-score scores and IoU values for each model and seismic event class at three window sizes: 80, 20, and 5 seconds. Best performance among models for each class and window size is indicated in bold.}
\label{tab:results_1}
\centering
\begin{tabular}{cc|ccc|ccc|ccc|ccc|ccc}
\hline
\hline
\multicolumn{2}{c|}{\textbf{Model}}                    & \multicolumn{3}{c|}{\textbf{UNet}}   & \multicolumn{3}{c|}{\textbf{UNet++}}          & \multicolumn{3}{c|}{\textbf{DeepLabV3+}}       & \multicolumn{3}{c|}{\textbf{SwinUNet}} & \multicolumn{3}{c}{\textbf{PhaseNet}} \\ \hline
\multicolumn{2}{c|}{\textbf{Window Size [s]}}          & 80            & 20            & 5    & 80            & 20            & 5             & 80            & 20            & 5              & 80    & 20             & 5             & 80    & 20            & 5             \\ \hline
\multicolumn{1}{c|}{\multirow{5}{*}{\textbf{F1}}} & VT & \textbf{0.91} & \textbf{0.84} & 0.60 & 0.88          & \textbf{0.84} & 0.69          & \textbf{0.91} & \textbf{0.84} & \textbf{0.70}  & 0.88  & 0.83           & 0.68          & 0.88  & \textbf{0.84} & 0.68          \\
\multicolumn{1}{c|}{}                             & LP & \textbf{0.90} & 0.67          & 0.33 & \textbf{0.90} & \textbf{0.68} & \textbf{0.61} & 0.89          & \textbf{0.68} & 0.54           & 0.88  & 0.63           & 0.50          & 0.00  & \textbf{0.68} & 0.57          \\
\multicolumn{1}{c|}{}                             & TR & \textbf{0.93} & 0.71          & 0.58 & 0.92          & 0.73          & \textbf{0.66} & 0.90          & \textbf{0.77} & 0.64           & 0.89  & 0.68           & 0.59          & 0.65  & 0.75          & 0.64          \\
\multicolumn{1}{c|}{}                             & AV & \textbf{0.87} & \textbf{0.72} & 0.49 & 0.86          & 0.70          & 0.56          & \textbf{0.87} & \textbf{0.72} & \textbf{0.57} & 0.86  & \textbf{0.72}  & 0.50          & 0.79  & 0.70          & 0.53          \\
\multicolumn{1}{c|}{}                             & IC & \textbf{0.93} & 0.89          & 0.60 & 0.92          & 0.88          & 0.73          & 0.91          & \textbf{0.92} & 0.71           & 0.90  & 0.91           & \textbf{0.74} & 0.87  & 0.89          & 0.68          \\ \hline
\multicolumn{2}{c|}{\textbf{Mean F1}}                  & \textbf{0.91} & 0.77          & 0.52 & 0.90          & 0.77          & \textbf{0.65} & 0.90          & \textbf{0.79} & 0.63           & 0.88  & 0.75           & 0.60          & 0.64  & 0.77          & 0.62          \\ \cline{1-2}
\multicolumn{2}{c|}{\textbf{Mean IoU}}                 & \textbf{0.88} & \textbf{0.87} & 0.90 & 0.87          & 0.86          & \textbf{0.94} & \textbf{0.88} & \textbf{0.87} & 0.93           & 0.86  & 0.85           & \textbf{0.94} & 0.82  & 0.85          & \textbf{0.94} \\ \hline
\end{tabular}
\end{table*}

\subsection{Noise Robustness}

\begin{figure}
\centering
\includegraphics[width=\linewidth]{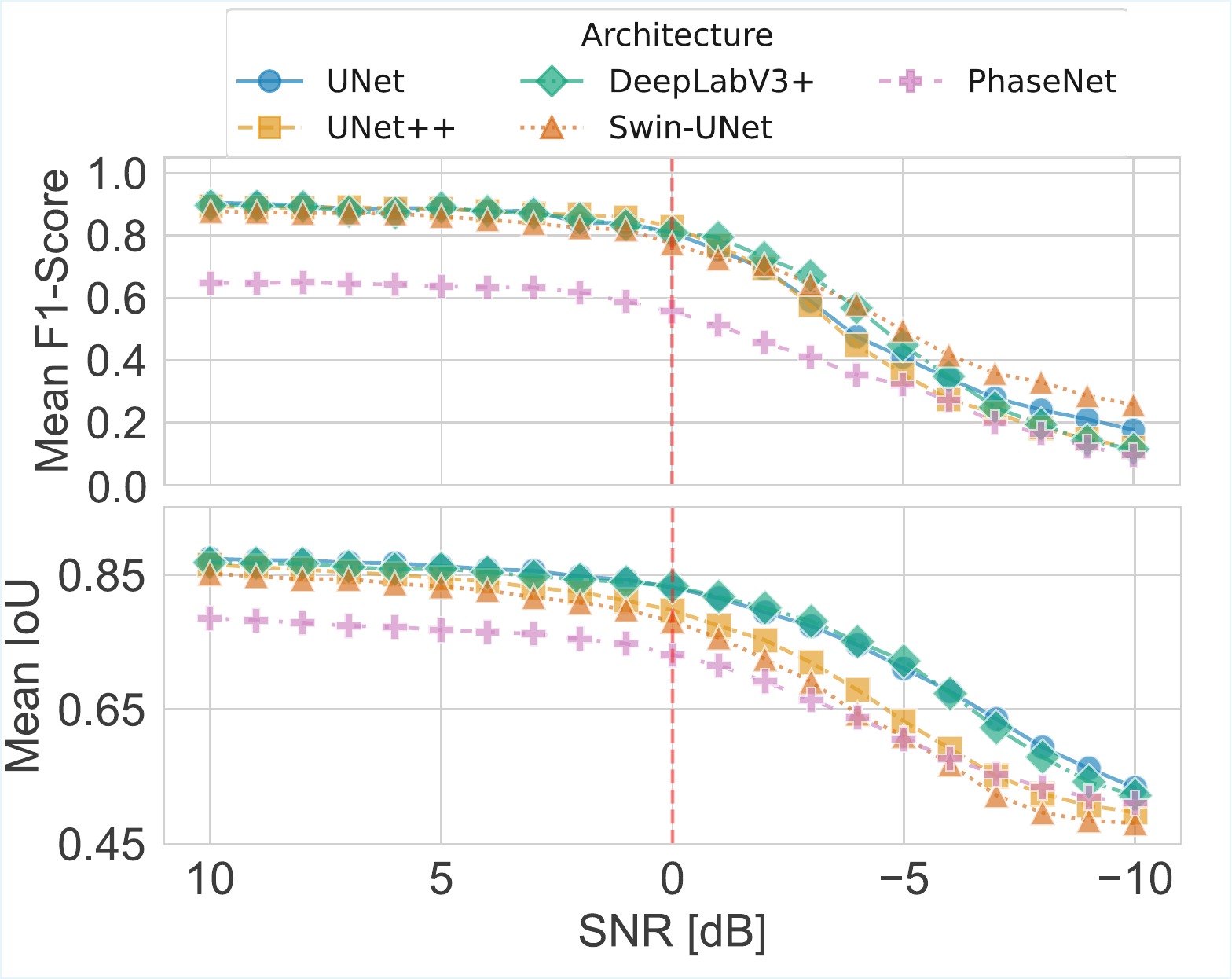}
\caption{Classification and detection performance through F1-score and IoU metric of the five models across SNR values ranging from 10 dB to -10 dB.}
\label{fig:res_noise}
\end{figure}

Figure~\ref{fig:res_noise} shows the performance of the five models across different SNR levels, measured by mean F1-score and IoU. The four 2-D models perform very similar in terms of F1-score, with Swin-UNet achieving the best performance under the lowest SNR conditions. When considering mean IoU, UNet and DeepLabV3+ perform best, with the difference becoming more noticeable at lower SNR levels. The only 1-D model, PhaseNet, consistently underperforms compared to the others in both F1-score and IoU.

Performance degradation is steeper for classification than for detection. For example, between SNR values of 10 and 0dB, the F1-scores of the models decreased by 10–17\%, while the average IoU dropped by only about 5\%. Moreover, at the lowest SNR level of –10dB, all IoU scores remained above 0.45, whereas F1-scores dropped below 0.2 for some models.

\subsection{Model Scalability}

\begin{figure*}[htbp]
\centering
\includegraphics[width=\linewidth]{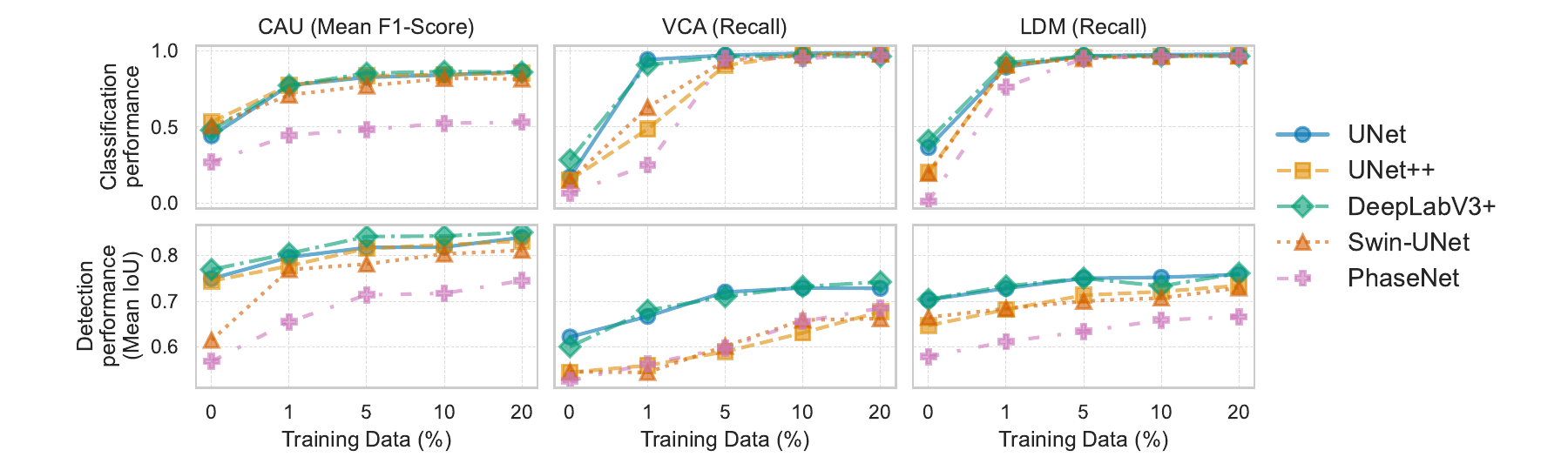}
  \caption{Scalability evaluation. Classification performance is shown in the top row and segmentation performance in the bottom row, as a function of the percentage of labeled target-volcano data used for mixed fine-tuning (0\% = zero-shot). For CAU (first column), classification is measured with mean F1-scores across three classes. For VCA and LDM (second and third columns), classification is reported using recall, since only a single class is available. Each curve corresponds to one of the five models.}
\label{fig:res_flexi}
\end{figure*}

Figure~\ref{fig:res_flexi} summarizes how the models adapt to the unseen CAU, VCA, and LDM datasets. In the zero-shot setting, the 2-D models (UNet, UNet++, DeepLabV3+, Swin-UNet) reach mid-range mean F1-score on CAU (0.44–0.53) with IoU 0.62–0.77, but show low VT recall on VCA/LDM (0.15–0.41) despite moderate IoU (0.55–0.70). With just 1\% of target labels under mixed fine-tuning with class completion, CAU mean F1-score jumps to $\sim$0.71–0.77 (IoU $\sim$0.77–0.80), VCA VT recall rises to 0.63–0.94 (IoU $\sim$0.55–0.68), and LDM VT recall to 0.76–0.92 (IoU $\sim$0.68–0.73). Increasing the fraction to 5–20\% yields diminishing returns: on CAU, mean F1-score stabilizes near 0.84–0.86 (best: DeepLabV3+ 0.863 at 10\%) with IoU around 0.81–0.85; on VCA and LDM, VT recall saturates at 0.96–0.98 for the 2-D models by 10–20\%, with detection IoU in the 0.66–0.76 range. LP remains the limiting class on CAU (LP F1-score improves from 0.09–0.43 in zero-shot to 0.68–0.78 at 10–20\%), while VT and TR are consistently high. PhaseNet improves markedly with data and reaches VT recall of 0.95–0.97 on VCA/LDM at 5–20\%, but its detection IoU (0.60–0.69) remains a few points below the best 2-D models (up to 0.74–0.76), and on CAU its mean F1-score lags throughout.

\subsection{Evaluation on Continuous Data}
\label{sec:cont_results}

\begin{figure}
\centering
\includegraphics[width=\linewidth]{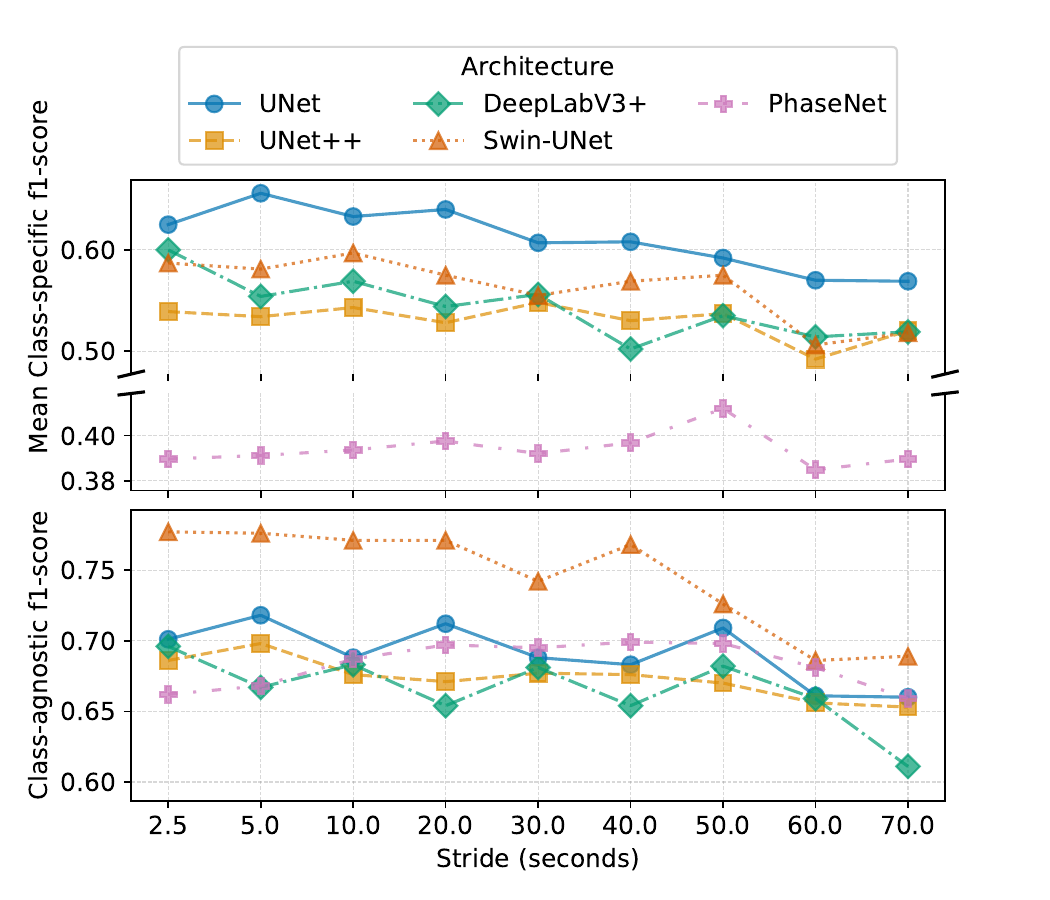}
\caption{Mean class-specific and class-agnostic F1-score as a function of the sliding-window stride (2.5–70~s). Class-specific F1-score (top) requires correct class and sufficient temporal overlap (IoP\,$\geq$\,0.5) with a reference event; class-agnostic F1-score (bottom) requires only sufficient overlap irrespective of class. The stride controls the model’s refresh rate (and alert latency).}
\label{fig:F1_vs_stride}
\end{figure}

Figure~\ref{fig:F1_vs_stride} shows how class-specific and class-agnostic F1-scores vary with the sliding-window stride. Most models benefit from shorter strides (higher refresh rate and overlap), with clear gains at 2.5–10~s, specially for UNet (detection and classification) and Swin-UNet (detection only). Performance degrades gradually as the stride grows beyond 30–40~s, consistent with sparser updates and fewer chances to capture event onsets.

PhaseNet scores considerably lower than the 2-D models in the detection-plus-classification setting across all strides, but closes the gap in the class-agnostic (detection-only) setting; in that context, Swin-UNet remains clearly superior.

Table~\ref{tab:stride_avg_metrics} aggregates precision, recall, and F1-scores, averaged across all nine strides. In the detection-plus-classification setting, UNet shows the best overall performance (precision, recall and F1-score), with the other 2-D models clustered behind and PhaseNet well below them. In the Class-Agnostic Detection setting, Swin-UNet attains the highest mean F1-score by a notable margin (0.75 vs.\ 0.69/0.68/0.67), yet UNet still achieves the best recall (0.78), reflecting a stronger coverage of cataloged events. Notably, PhaseNet ranks third in terms of F1-score under detection-only.

\begin{table}[t]
\centering
\caption{Mean ($\pm$ std) precision, recall, and F1-score across the nine stride lengths (2.5–70 s). Detection \& Classification reports the average metrics accross classes, while class-agnostic detection performance considers a correct prediction as long as there is an overlap with a reference event.}
\label{tab:stride_avg_metrics}
\begin{tabular}{r|c|ccc}
\hline\hline
                       & \multicolumn{1}{c|}{\textbf{Model}} & \textbf{Precision}                       & \textbf{Recall}                 & \textbf{F1-score}                        \\ \hline
\multirow{5}{*}{\begin{tabular}[c]{@{}c@{}}Detection \& \\Classification\end{tabular}} 
                       & \textbf{UNet}                       & \textbf{0.59$\pm$0.02}                  & \textbf{0.68$\pm$0.02}        & \textbf{0.61$\pm$0.02}                  \\
                       & \textbf{UNet++}                     & 0.55$\pm$0.01                            & 0.55$\pm$0.01                  & 0.53$\pm$0.01                            \\
                       & \textbf{DeepLabV3+}                 & 0.58$\pm$0.02                            & 0.55$\pm$0.02                  & 0.54$\pm$0.02                            \\
                       & \textbf{SwinUNet}                   & 0.56$\pm$0.03                            & 0.58$\pm$0.01                  & 0.56$\pm$0.02                            \\
                       & \textbf{PhaseNet}                   & 0.40$\pm$0.01                            & 0.50$\pm$0.01                  & 0.39$\pm$0.01                            \\ \hline
\multirow{5}{*}{\begin{tabular}[c]{@{}c@{}}Class-Agnostic\\Detection\end{tabular}} 
                       & \textbf{UNet}                       & 0.62$\pm$0.02                            & \textbf{0.78$\pm$0.01}        & 0.69$\pm$0.01                            \\
                       & \textbf{UNet++}                     & 0.64$\pm$0.01                            & 0.71$\pm$0.02                  & 0.67$\pm$0.01                            \\
                       & \textbf{DeepLabV3+}                 & 0.65$\pm$0.02                            & 0.68$\pm$0.02                  & 0.67$\pm$0.02                            \\
                       & \textbf{SwinUNet}                   & \textbf{0.75$\pm$0.04}                   & 0.75$\pm$0.01                  & \textbf{0.75$\pm$0.02}                   \\
                       & \textbf{PhaseNet}                   & 0.68$\pm$0.01                            & 0.69$\pm$0.01                  & 0.68$\pm$0.01                            \\ \hline
\end{tabular}
\end{table}

\section{Discussion}

\textbf{Effect of Window Size:}  
To gauge the impact of temporal context, we evaluated all models at multiple window lengths. Performance dropped sharply when the window was reduced from 80 s, indicating that broad context is essential for reliable event discrimination. PhaseNet—the 1-D baseline—performs much closer to the 2-D models in terms of detection than in terms of classification, reflecting its design for binary earthquake detection and its limitations in multi-class tasks.

Our maximum window length (80\,s) was constrained by database curation and computational limitations. However, given the results, we believe that longer windows could provide better performance and robustness in general and should be explored in future works.

\textbf{Model Comparison:} 
The UNet architecture proved to be the best model overall. This superiority may be due to the fact that more complex architectures, like UNet++, DeepLabV3+, and SwinUNet, are designed for image segmentation tasks where abstract features and more complex embeddings represent important advantages. In seismic signal analysis, however, event differentiation occurs largely at the time–frequency level, for which standard convolutional layers appear sufficient. PhaseNet consistently underperformed compared to the 2-D models, matching their accuracy only on single-event datasets (e.g., VCA and LDM) or when only detection is evaluated (see Figure \ref{fig:F1_vs_stride}). This further underscores its design limitations: although effective for binary earthquake detection (demonstrated, for instance, in \cite{Kim2023}, where PhaseNet was successfully adapted for phase picking of VT events at Hakone volcano) it lacks the architectural flexibility required for multi-class event recognition tasks. Notably, DeepLabV3+ packs just 3.2 million parameters—fewer than PhaseNet’s 4.3 million—yet it systematically outperformed PhaseNet and ranked second overall in many of our evaluations.

\textbf{Adaptability and Robustness:}
The models showed strong noise robustness, particularly in terms of detection, and adapted to data from other volcanoes after minimal retraining with only a small amount of new samples. Under cross-volcano transfer the IoU remains moderate in zero-shot and improves with a small amount of target data. By contrast, classification is more sensitive to dataset changes, as class-defining seismic patterns change among volcanoes. This is why the models tend to detect events before they can reliably differentiate among classes. With mixed fine-tuning and class completion, 1–5\% labeled target data is sufficient to re-align class boundaries, producing sharp gains in classification while preserving the detection robustness already present in zero-shot. Although detection performance on the VCA and LDM datasets did not reach the level obtained on the original NChVC dataset, given the consistently high classification scores, we attribute this shortfall to differences at the moment of including seismic codas (i.e. correctly choosing end times for VT events) rather than faulty detections from the models.

\textbf{Continuous-Data Performance:}  
Under the operational setting in Fig.~\ref{fig:method_0_diagram}, and applying a fixed amplitude threshold on detected segments (RSAM$_{net}$), the 2-D models attain class-specific (detection + classification) F1-scores in the 0.53–0.61 range across strides, and class-agnostic (detection-only) F1-scores in the 0.67–0.75 range. The baseline PhaseNet remains clearly worse once classification is required, but its detection-only F1-score (0.68 on average) rises into the same band as most 2-D models. This highlights PhaseNet’s suitability for single-class detection while underscoring the 2-D models’ advantage for multi-class event labeling. If sensitivity is prioritized, UNet delivers the highest recall in both class-aware and class-agnostic settings (0.68 and 0.78, respectively).

Processing the full 10-hour trace took as little as 4 seconds with UNet at a 70-second stride and up to 5 minutes with SwinUNet at a 2.5-second stride on an NVIDIA RTX 3060 GPU (see the continuous-data demonstration in our repository, Sec.~\ref{section:code_availability}). On average, per-window processing times ranged from 7 to 23~ms, with UNet being the fastest and SwinUNet the slowest. The refresh rate is primarily determined by the stride, as window throughput contributes only marginally (on the order of milliseconds). As shown in Fig.~\ref{fig:F1_vs_stride}, shorter strides (2.5--10~s) consistently yield performance gains across models. We therefore regard the system as a real-time detector, since the refresh rate can be controlled through the stride and, in principle, reduced to a single sample (0.01~s).

Because observatories prioritize not missing hazardous events, we treat recall as the primary safety metric; class-agnostic results confirm solid detection capability in continuous processing. In addition, models can be paired with a simple uncertainty proxy based on activation proportions within each detected segment (see the window-level demonstration in Sec.~\ref{section:code_availability}). Increasing overlap (shorter strides) not only improves accuracy but also provides multiple forward passes over the same signal patch, enabling more stable uncertainty estimates.

\textbf{1-D vs 2-D segmentation:}
2-D models clearly outperform the 1-D PhaseNet baseline across our tests when classification is taken into account, while in detection only context (only VT events or class-agnostic detection in continuous data) PhaseNet performs very similar to them. One possible explanation is that stacking the eight synchronous traces into a two-dimensional map endows the network with genuine spatio-temporal equivariance: a small 2×2 or 3×3 filter can translate both in time and across stations, intuitively picking up patterns like the phase-shift between stations, polarity flips (first‐motion sign reversals between adjacent stations), or amplitude gradients (systematic decay of peak amplitude across the array due to geometric spreading and attenuation). By contrast, a purely 1-D UNet like PhaseNet (despite being widened to 4.2 M parameters) necessarily collapses all eight channels into a single feature vector at each time step, possibly losing relevant station information at deeper levels of processing. This could introduce redundancy and limit generalization when multiple volcanic classes with subtle, station-dependent morphologies must be distinguished. That PhaseNet matches the 2-D models on single-class VT datasets may reflect that, in a binary detection task, cross-station cues matter less; once class competition increases, however, lacking a built-in notion of station locality appears to undermine its segmentation accuracy. In contrast, the 2-D architectures appear to consistently fuse fine temporal detail with local station context at every scale, which may explain their superior multi-class performance.

\section{Conclusions}

We have developed a novel approach for seismic event recognition using 2-D semantic-segmentation models, and used it to conduct experiments on 24,493 labeled windows across multiple volcanoes, as well as on a 10-hour continuous trace. Among the four architectures evaluated, UNet consistently delivered the best performance, strong noise robustness, easy adaptation to unseen data, and solid results on continuous streams. It achieved a mean F1-score of 0.91 and an IoU of 0.88 for window-level recognition, as well as a mean class-aware recall of 0.68 and a class-agnostic recall of 0.78 on continuous data. To the best of our knowledge, our framework is the first to integrate information from multiple seismic stations to perform both detection and classification for multi-class volcano-seismic events. Moreover, it requires only lightweight preprocessing and can process a 10-hour multichannel trace within minutes. The 2-D representation we propose opens new avenues for handling continuous seismic records, incorporating additional variables of interest, and leveraging the extensive research in semantic segmentation. This framework can significantly enhance volcano-monitoring operations by automating repetitive detection and classification tasks, allowing seismologists to focus on deeper analyses of seismic phenomena and their links to volcanic processes. However, we would like to emphasize that, although real-time event recognition is a powerful and evolving tool, it must not replace human expertise. Volcanoes are hazardous and unpredictable systems, and reliable monitoring is critical for public safety. Automated methods should be adopted gradually and with caution, as they remain prone to errors and require expert oversight to ensure accurate and accountable alert decisions.

\subsection*{Limitations and Future Work}

As noted in Section~\ref{sec:window_detection}, our models can inherently localize overlapping events and produce station-specific detections. To simplify the present evaluation, we assumed no overlaps and analyzed only time-based segmentation. Because event overlap is common during periods of heightened volcanic activity, and stations can be more or less informative at each moment in time, future work should exploit these native capabilities. Doing so will require a newly curated dataset that explicitly labels overlaps and a revised strategy for evaluating detections.

Performance improved as window length increased. However, using longer windows poses higher computational costs and requires a different evaluation scheme from the one proposed here, as they naturally incorporate multiple events and the amount of single-event windows diminishes the larger the window size is. Future work focused on curating longer windows and proposing a new preprocessing (e.g. subsampling windows) and evaluation scheme could substantially boost model performance and make better use of existing databases.

Our \emph{Patch Stacking} approach converts multi-station 1-D traces into single-channel 2-D images, leveraging the already matured deep semantic segmentation literature. This framework can be extended to accept multi-channel inputs with hand-crafted features—such as signal power, moving averages, entropy, or autocorrelation—to further enhance performance. Additionally, output-refinement techniques—such as conditional random fields, temporal smoothing filters, or alternative loss functions—could yield further performance gains.

A natural next step is to isolate the spatial inductive bias of the stacking process itself. For example, one could compare how 1-D versus 2-D models handle synthetic waveforms with controlled inter-station time‐shifts, polarity reversals and amplitude decay—patterns that 2-D folding should capture more naturally. Analyzing filter activations and feature reuse across stations, or testing on novel station geometries, would reveal whether the 2-D layout intrinsically embeds cross-sensor equivariance, rather than simply benefiting from incidental differences in capacity or training.
  
False positives are still a challenge in our approach. The fact that most of them are pure noise patches, indicates that our trace normalization procedure (Section \ref{sec:data_prep}) could be improved to deal with event amplitude in a more intelligent manner. Another road is to curate representative noise-only windows to use during training, although these require careful revision and a clear definition of what type of small-amplitude events can be ignored.

Because the models show complementary strengths, one possible extension would be to combine them through ensemble or voting strategies. However, running several large networks in parallel would raise inference latency and larger hardware demands, which is problematic for real-time monitoring. Future work could explore more efficient ensemble designs, such as lightweight model combinations or knowledge distillation. This could balance accuracy improvements with the computational constraints of operational monitoring.

Automatic volcano monitoring requires not only accurate detection and classification but also reliable uncertainty estimates. Although our repository demonstrates a coarse proxy based on activation proportions (Section~\ref{section:code_availability}), developing computationally efficient, fine-grained uncertainty and anomaly measures remains a critical goal.

Finally, analysis of layer-activation maps and ablation studies can reveal important details on how the models process their inputs, improving interpretability and guiding targeted model refinements.

\section*{Acknowledgments}
The authors gratefully acknowledge OVDAS-Sernageomin for providing the data and expert insights, the CIVUR-39° N°FRO2193/CIV23-0001 project of the Research Directorate at UFRO, FONDEF IT2310036 project, and FONDECYT N°11230289 project for the financial support that enabled the development of the database and the models used in this study.

\section{Data Availability}

Data supporting the findings of this study are openly available in Zenodo at \url{https://doi.org/10.5281/zenodo.15384923}. It includes seismic signals processed according to the procedure described in section \ref{section:database}. The weights of the four main models we developed are also available at \url{https://doi.org/10.5281/zenodo.15098817}.

\section{Code availability}
\label{section:code_availability}
Contact: camilo.espinosa@ug.uchile.cl. Hardware requirements: Dedicated GPU with over 1GB capacity and CUDA compatibility. Program language: Python. The source codes and demonstration scripts are available at: \url{https://github.com/camilo-espinosa/volcano-seismic-segmentation}

\bibliographystyle{IEEEtran}
\bibliography{bibliography}
\newpage
\section{Biography Section}

\begin{IEEEbiography}[{\includegraphics
[width=1in,height=1.25in,clip,
keepaspectratio]{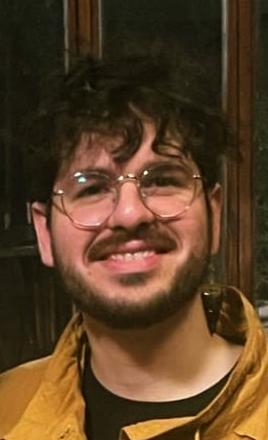}}]
{Camilo Espinosa-Curilem} 
received his Master’s degree in Engineering Sciences, with a focus on Electrical Engineering, from Universidad de Chile in 2023. He is currently a Development and Analysis Engineer at the Advanced Mining Technologies Center (AMTC). His research interests include deep learning, artificial intelligence, information theory, and neuroscience.
\end{IEEEbiography}
\begin{IEEEbiography}[{\includegraphics
[width=1in,height=1.25in,clip,
keepaspectratio]{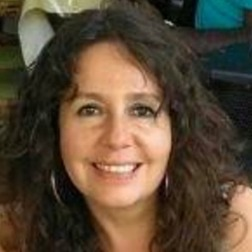}}]
{Millaray Curilem Saldías} 
earned her Computer Engineering degree in Havana, Cuba (1991), later recognized by Universidad de Chile as Civil Engineer in Computing. She is Dr. in Electrical Engineering from the Federal University of Santa Catarina, Brazil (2002) and has two postdoctoral positions in Chile (2007) and Brazil (2017). Since 1994, she has worked at Universidad de La Frontera in Temuco, Chile, where she is a Full Professor in the Department of Electrical Engineering. Her research focuses on computational intelligence for pattern recognition and signal processing. She is a member of the IEEE Computational Intelligence Society since 2007.
\end{IEEEbiography}
\begin{IEEEbiography}[{\includegraphics
[width=1in,height=1.25in,clip,
keepaspectratio]{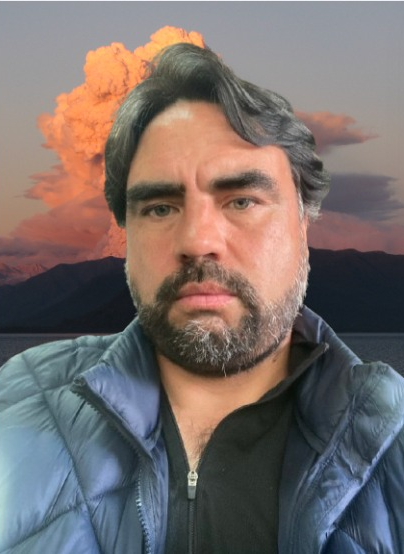}}]
{Daniel Basualto Alarcón} 
is a geologist and volcanologist, with a Ph.D. in Geological Sciences from Universidad de Concepción. His research focuses on volcanic systems, seismic activity, and the development of real-time geophysical monitoring networks. He has extensive experience in volcanic crisis management in South America, especially in Chile and Ecuador, and has contributed to numerous high-impact scientific publications. Currently, he is a professor at Universidad de La Frontera, leading research projects funded by FONDECYT, in addition to participating in technology transfer programs and fieldwork in extreme environments.
\end{IEEEbiography}

\end{document}